\documentclass[nonacm, sigconf]{acmart}

\author{Mohammad Mehdi Rastikerdar}
\email{mrastikerdar@cs.umass.edu}
\affiliation{\institution{University of Massachusetts Amherst}
\country{United States}}

\author{Jin Huang}
\email{jinhuang@cs.umass.edu}
\affiliation{\institution{University of Massachusetts Amherst}\country{United States}}

\author{Hui Guan}
\email{huiguan@cs.umass.edu}
\affiliation{\institution{University of Massachusetts Amherst}\country{United States}}

\author{Deepak Ganesan}
\email{dganesan@cs.umass.edu}
\affiliation{\institution{University of Massachusetts Amherst}\country{United States}}
\usepackage{comment}
\usepackage{amsmath}
\usepackage{wrapfig}
\usepackage{subcaption}
\usepackage{setspace}
\usepackage{listings}
\usepackage{graphicx}
\usepackage{graphics}
\usepackage{amsmath}
\usepackage{pgfgantt}
\usepackage{pgfplots}
\usepackage{tikz}
\usepackage{amsfonts}
\usepackage{algorithmic}
\usepackage{multirow}
\usepackage{ulem}
\usepackage{color}
\usepackage{array}
\usepackage{enumitem}
\usepackage{hyperref}

\newcommand{\dg}[1]{{\color{teal}\textit{ \{DG: #1\}}}}

\newcommand{\mypar}[1]{\vspace{0.15cm} \noindent {\textbf{#1.}\/}}
\newcommand{\pjn}{WildFiT}

\begin{document}


\title{\pjn: Autonomous In-situ Model Adaptation for Resource-Constrained IoT Systems}

\begin{abstract}
Resource-constrained IoT devices increasingly rely on deep learning models, however, these models experience significant accuracy drops due to domain shifts when encountering variations in lighting, weather, and seasonal conditions. While cloud-based retraining can address this issue, many IoT deployments operate with limited connectivity and energy constraints, making traditional fine-tuning approaches impractical. We explore this challenge through the lens of wildlife ecology, where camera traps must maintain accurate species classification across changing seasons, weather, and habitats without reliable connectivity. We introduce \pjn{}, an autonomous in-situ adaptation framework that leverages the key insight that background scenes change more frequently than the visual characteristics of monitored species. \pjn{} combines background-aware synthesis to generate training samples on-device with drift-aware fine-tuning that triggers model updates only when necessary to conserve resources. 
Our background-aware synthesis surpasses efficient baselines by 7.3\% and diffusion models by 3.0\% while being orders of magnitude faster, our drift-aware fine-tuning achieves Pareto optimality with 50\% fewer updates and 1.5\% higher accuracy, and the end-to-end system outperforms domain adaptation approaches by 20-35\% while consuming only 11.2 Wh over 37 days—enabling battery-powered deployment.

\end{abstract}

\maketitle

\section{Introduction}


Resource-constrained IoT devices increasingly rely on deep learning models to enable intelligent applications in remote environments. However, maintaining model accuracy in real-world deployments remains challenging due to domain shifts i.e. when the statistical properties of incoming data differ significantly from the training data~\cite{blanchard2011generalizing, zhou2022domain}. These shifts are particularly pronounced in outdoor IoT applications where environmental conditions like lighting, weather patterns, and seasonal changes can dramatically alter the visual characteristics that models rely on for inference.

This domain shift limits the deployment of ML models on IoT devices. While larger, more robust models can help maintain accuracy across domains, their computational demands often exceed the capabilities of resource-constrained devices. Recent advances in domain generalization through techniques like domain alignment~\cite{muandet2013domain}, meta-learning~\cite{li2018learning}, and ensemble methods~\cite{arpit2022ensemble, yeo2021robustness}) typically require increased model capacity to learn and represent broader feature sets that distinguish different domains. This presents a fundamental tension between achieving high generalization performance and ensuring efficient inference on IoT platforms.

We explore this challenge through the lens of wildlife ecology, where camera traps serve as critical tools for monitoring animal behavior and populations~\cite{lila_datasets,glover2019camera,falzon2019classifyme,price2016animalfinder,yousif2019animal,ahumada2020wildlife}. These motion-triggered cameras are often deployed in remote areas with limited access to power and network connectivity, where relying on cloud processing can lead to significant delays and costs (estimated at weeks to months between capture and analysis and \$2.15M for field visits and data retrieval ~\cite{smith2024man}). This has driven the development of \textit{on-device animal detection and classification systems}~\cite{tabak2019machine, beery2018recognition, lila_datasets} that process images locally, enabling immediate species identification despite limited network connectivity. 

However, these on-device models face unique challenges that make them an ideal testbed for studying domain adaptation in resource-constrained settings. Specifically, the deployment environments introduce both spatial shifts arising from variations in lighting, vegetation, and terrain across locations, and temporal shifts due to weather patterns and seasonal changes - challenges that can severely degrade model accuracy but must be addressed within the computational constraints of IoT devices.

\mypar{Limitation of Existing Approaches} For resource-constrained IoT devices, model fine-tuning offers a promising approach to address domain shifts without requiring the computational overhead of larger, more complex models like those used for domain generalization~\cite{muandet2013domain, zhou2022domain}. However, enabling efficient and effective fine-tuning in remote deployments presents several key challenges.

First, fine-tuning approaches fundamentally rely on collecting representative data from the target domain which is often difficult to obtain. In domains like wildlife monitoring, target domain data is inherently sparse - animals appear infrequently and obtaining ground truth labels requires manual verification~\cite{beery2018recognition}, making it impractical to build comprehensive datasets for new environments. Even unsupervised approaches that leverage pseudo-labels struggle, as the poor model performance after domain shifts leads to noisy and unreliable labels~\cite{blanchard2011generalizing}. This challenge extends to many IoT applications where events of interest are infrequent and labeling requires domain expertise.

Second, fine-tuning must be proactive to prevent accuracy degradation, yet determining when to trigger updates remains an open challenge. The ideal approach would adapt models before significant performance drops occur, but this requires predicting domain shifts without access to labeled target domain data. This creates a circular dependency - we need target domain data to detect shifts and trigger fine-tuning, but we want to fine-tune before collecting significant target domain data to maintain high accuracy. In wildlife monitoring, for instance, by the time enough animal images are collected to validate a domain shift, the model may have already experienced substantial accuracy degradation~\cite{tabak2019machine, lila_datasets}.

These limitations highlight the need for new approaches that can: (1) enable fine-tuning without relying on extensive target domain data collection, and (2) proactively trigger model updates based on early indicators of domain shift. Crucially, any solution must also work \textit{within the computational and energy constraints} of IoT devices deployed in remote environments with limited connectivity~\cite{beery2018recognition, lila_datasets}.

\mypar{Our Approach} In this work, we introduce \pjn{}, an \textit{in-situ fine-tuning} system that enables camera traps to autonomously adapt to domain shifts through purely on-device operations, eliminating the need for external connectivity or resources.

The key insight driving \pjn{} is that we can leverage easily observable background changes to anticipate and address domain shifts in camera trap deployments. There are two types of domain shifts that occur in camera traps: environmental changes (like terrain, vegetation, lighting, and weather) and shifts in animal distributions. While animal appearances are infrequent and unpredictable, background scenes from the static camera provide a continuous stream of information about environmental changes. By combining these background images with a compact repository of animal objects from the source domain, \pjn{} can synthesize high-fidelity training data that captures current domain conditions without waiting to collect actual animal images.

This synthesis capability enables a novel proactive adaptation approach. Rather than waiting to observe accuracy degradation on real animal images, \pjn{} continuously generates and evaluates synthetic images that reflect current conditions. When these evaluations predict potential performance drops, \pjn{} can immediately fine-tune the model using synthesized training data. This approach breaks the traditional dependency on collecting target domain data, allowing camera traps to adapt rapidly to changing conditions while operating autonomously in remote environments.

\mypar{Technical Challenges} Implementing background-aware model adaptation presents two fundamental challenges. The first challenge lies in developing an efficient yet high-quality data synthesis method. While collecting background images is straightforward, generating realistic training samples by blending animals into these backgrounds is non-trivial on resource-constrained devices. Existing approaches fall on opposite ends of the compute-quality spectrum: diffusion models can generate photorealistic compositions but require significant computational resources that exceed IoT capabilities, while traditional augmentation methods like image blending and mixing are computationally efficient but produce low-quality samples that fail to capture the nuanced interactions between animals and their environments.

The second challenge involves determining the optimal timing for model updates. Without direct access to labeled target domain data, identifying when fine-tuning is necessary becomes complex. Triggering updates too late (false negatives) leads to prolonged periods of degraded performance, while unnecessary updates (false positives) waste precious computational resources on devices with limited power budgets. This creates a need for robust criteria that can predict accuracy drops without relying on ground truth labels from the target domain.

To address these challenges, \pjn{} introduces two key innovations. First, \textit{background-aware data synthesis} enables efficient generation of high-quality training data directly on IoT devices by intelligently integrating source domain animals into current backgrounds. Unlike computationally expensive approaches like diffusion models or simplistic blending techniques, our method preserves crucial visual relationships between animals and their environments while remaining lightweight enough for resource-constrained devices.

Second, \textit{drift-aware fine-tuning} leverages these synthesized images to enable a powerful new approach to model adaptation. Whenever a potential drift is detected, drift-aware fine-tuning first invokes a drift validation module to evaluate model accuracy using synthesized images that reflect current environmental conditions and species distributions. It triggers fine-tuning only when drift validation confirms accuracy drops. This allows camera traps to anticipate and address accuracy drops before they impact wildlife monitoring, while carefully managing computational resources by triggering fine-tuning only when necessary.

We conducted extensive evaluations across three camera trap datasets and four platforms. Our results show that 
\begin{itemize}
\item \pjn{}'s \textit{background-aware data synthesis} surpasses other computationally efficient approaches~\cite{yun2019cutmix, zhang2017mixup} by 7.3\% and diffusion model-based synthesis~\cite{objectstitch, controlcom} by 3.0\% in model accuracy, while being several orders of magnitude faster (10-150 ms vs 300-600 seconds for a batch of 32 images). 
\item \pjn{}'s \textit{drift-aware fine-tuning} achieves Pareto optimality in terms of fine-tuning frequency and classification accuracy under drifts. It achieves up to 1.5\% higher accuracy and requiring 50\% fewer fine-tuning rounds compared to periodic fine-tuning. 
\item End-to-end results show that \pjn{} significantly outperforms several domain adaptation approaches by 20-35\%, even surpassing methods that utilize ground-truth labels of animals in the new location for fine-tuning. We also show that a single fine-tuning iteration on a batch of 32 images takes 2-68 seconds depending on the IoT platform, demonstrating its practicality for real-world deployments. 
\item The drift validation module within \pjn{} significantly reduces both energy consumption and computational latency by preventing unnecessary fine-tuning iterations, as it is up to 160$\times$ faster than one round of fine-tuning and achieves up to 200$\times$ lower energy usage across different power modes on the Jetson Orin Nano Super platform.
\item The end-to-end \pjn{} pipeline is highly energy-efficient and consumes merely 11.2 Wh for processing a real stream of camera trap images spanning 37 days—a consumption level supported by three alkaline AA batteries.
\end{itemize}


\section{Background and Motivation}
\label{sec:background}
This section introduces camera trap applications and elaborates on the domain shift problems in this space.



\mypar{Camera Trap Applications} Camera traps are motion-triggered cameras that are widely used methods for ecological monitoring in remote, often inaccessible locations~\cite{bruce2024large}. While cloud-based platforms can leverage powerful AI algorithms for species identification~\cite{ahumada2020wildlife}, the high costs of field visits and data retrieval (estimated at \$2.15M for a typical monitoring program~\cite{smith2024man}) have driven increasing interest in on-device processing. These devices leverage machine learning (ML) models for automated species identification, movement tracking, and rare species detection directly on the device~\cite{falzon2019classifyme}. Processing data locally through \textit{on-device inference} offers three key benefits: it eliminates the substantial costs and carbon footprint associated with frequent field visits~\cite{smith2024man}, enables real-time responses to time-critical events like invasive species detection~\cite{ramsey2011quantifying}, and ensures continuous monitoring even in areas with unreliable network connectivity~\cite{glover2019camera, rastikerdar2024cactus}.

\mypar{The Domain Shift Problem}
Domain shift occurs when the data encountered during deployment (target domain) differs significantly from that used to train the model (source domain). This mismatch between training and real-world conditions can severely degrade a model's performance, even if it achieved high accuracy during training. For example, a wildlife classification model trained on images from sunny days may struggle when processing images captured in rainy conditions, as the visual features it learned no longer match the new environmental context.

\begin{figure}[t]
    \centering
    \includegraphics[width=0.95\linewidth]{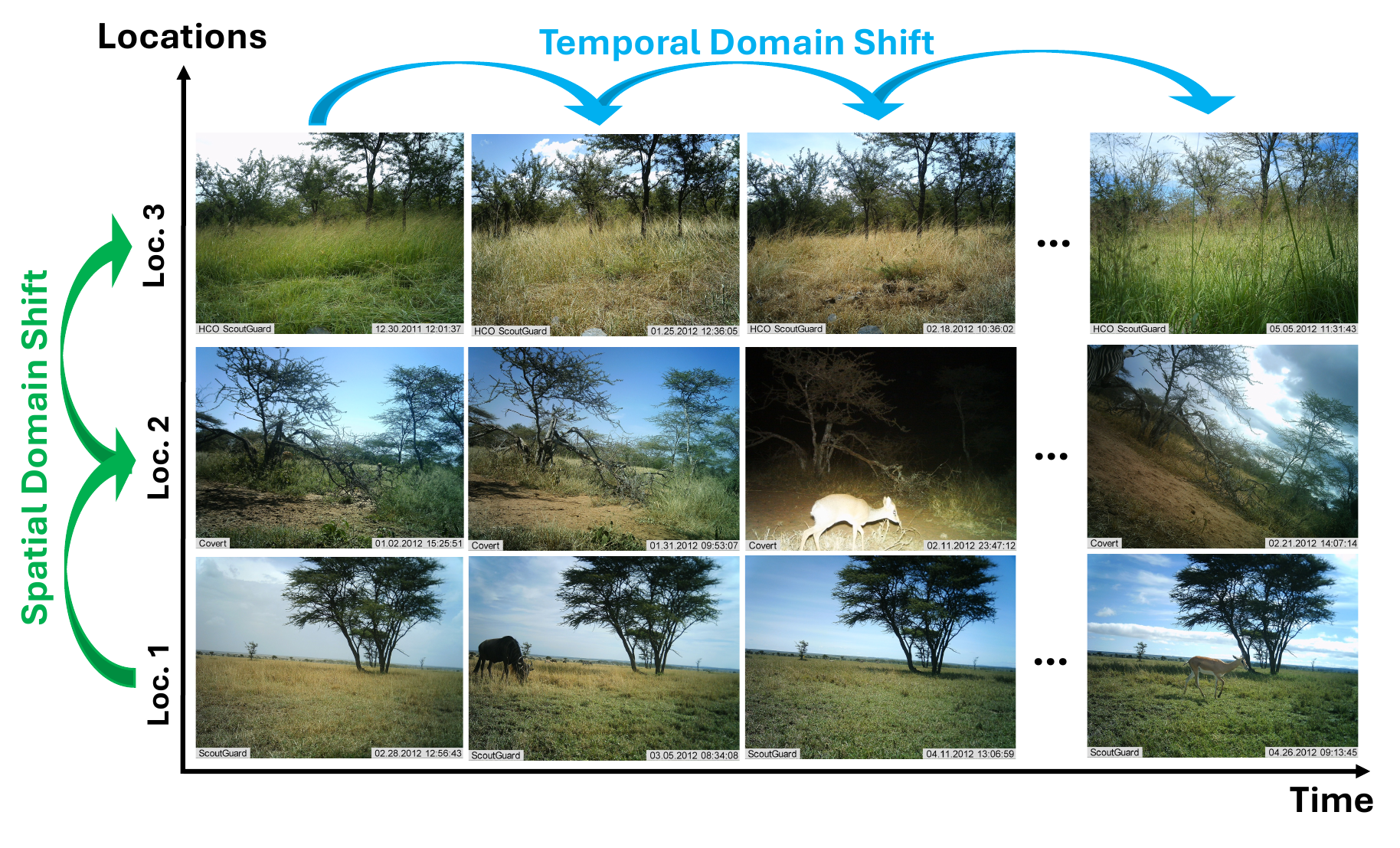}
    \vspace{-.4cm}
    \caption{Spatial and temporal domain shifts in camera trap applications. Specifically, Location 2's camera position shifted over time while Location 1 and 3's backgrounds show seasonal changes. As we show in Table~\ref{tab:generalization_problem}, these domain shift can cause an 9\% - 60\% drop in wildlife recognition accuracy.}
    \label{fig:domain_shift}
    \vspace{-0.3cm}
\end{figure}

In camera trap applications, each new deployment site introduces unique environmental conditions -- such as lighting, vegetation, terrain, and local wildlife -- that lead to \textit{spatial domain shift}. Furthermore, weather patterns, seasonal variations, and other time-based changes contribute to a \textit{temporal domain shift}, which alters the appearance of captured images over time. Figure~\ref{fig:domain_shift} illustrates the two types of domain shift in camera trap settings.

\begin{table}[t]
    \centering
    \small 
    \setlength{\tabcolsep}{0.08cm} 
    \begin{tabular}{c!{\vrule width 1.5pt}c|c|c!{\vrule width 1.5pt}c|c|c!{\vrule width 1.5pt}}
        \hline\hline
        \textbf{Locs.} & \textbf{Before} & \textbf{After} & \textbf{Acc.} & \textbf{Before} & \textbf{After} & \textbf{Acc.} \\
         & \textbf{Spatial} & \textbf{Spatial} & \textbf{Drop} & \textbf{Temporal} & \textbf{Temporal} & \textbf{Drop} \\
         & \textbf{Shift} & \textbf{Shift} & & \textbf{Shift} & \textbf{Shift} &  \\
        \hline
        1 & 78.8 & 66.4 & 12.4 & 80.6 & 52.1 & 28.5 \\
        2 & 78.8 & 19.3 & 59.5 & 67.9 & 59.2 & 8.7 \\
        3 & 78.8 & 66.3 & 12.5 & 87.9 & 63.8 & 24.1 \\
        \hline\hline
    \end{tabular}
    \caption{The wildlife classification model accuracy before
and after spatial and temporal domain shifts on 3 test locations (R06, R09, and T11) of Serengeti S4~\cite{snapshotserengeti}}
    \label{tab:generalization_problem}
    \vspace{-0.3in}
\end{table}

\mypar{Domain shifts cause significant performance drop} Table~\ref{tab:generalization_problem} shows the performance gap caused by spatial and temporal domain shifts when using EfficientNet-B0, a small classification model that is typically used on IoT platforms. To demonstrate the impact of \textit{spatial domain shift}, we trained a wildlife classification model using images from 154 camera trap locations in the Serengeti S4 dataset (details in \S\ref{sec:settings}). The "Before Spatial Shift" accuracy is measured on samples from the same 154 locations, excluding those used in training. Testing the model on three new, unseen locations reveals that spatial domain shifts can cause accuracy drops of 12\%--60\%.

We use the same three locations to highlight the impact of \textit{temporal domain shift}. The data from each location is split into three equal chunks. The model is trained using data from the mentioned 154 locations, along with 20\% of the beginning of the first chunk. "Before Temporal Shift" accuracy is computed on the remaining 80\% of the first chunk. To assess "After Temporal Shift" accuracy, we test on the last chunk, which likely has the greatest temporal shift, resulting in a 9-28\% reduction in accuracy.

We see that both these shifts can have a substantial impact on performance. A model trained on data from one location may perform poorly when deployed in a different environment with unfamiliar backgrounds or lighting conditions. In addition, the specific species of animals that are frequent in an area can vary across locations and seasons, further complicating the classification task.

\mypar{Fine-tuning restores model performance}
A wide range of domain generalization methods, including domain alignment~\cite{muandet2013domain}, meta-learning~\cite{li2018learning}, 
and ensemble learning~\cite{arpit2022ensemble, yeo2021robustness}), have proven effective in mitigating the effects of domain shift. 
They typically necessitate larger model architectures, as the extensive variations inherent in the data require increased capacity (e.g., more layers and parameters) to adequately learn and represent the broader set of distinguishing features across domains. Nonetheless, our findings indicate that fine-tuning remains a highly effective strategy for restoring model performance, irrespective of the model complexity.

\begin{figure}[t]
    \centering
    \begin{subfigure}[t]{0.23\textwidth}
        \centering
        \includegraphics[height=1.2in]{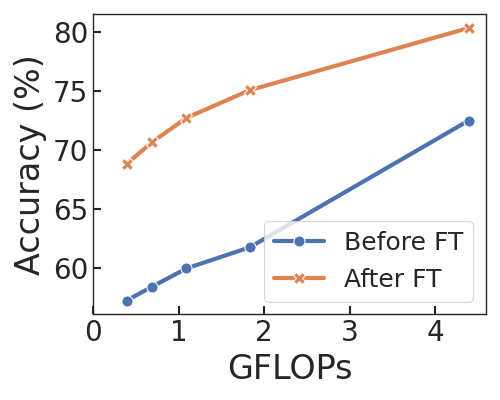}
        \vspace{-.15cm}
        \caption{A new location.}
    \end{subfigure}
    ~
    \begin{subfigure}[t]{0.23\textwidth}
        \centering
        \includegraphics[height=1.2in]{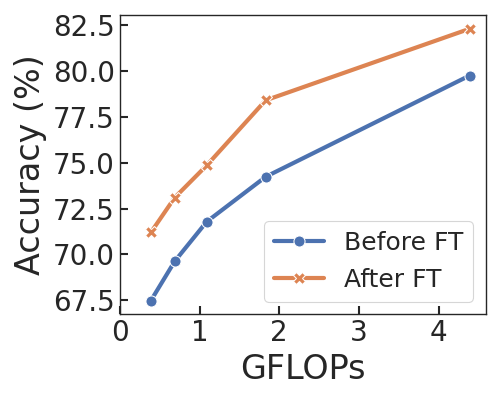}
        \vspace{-.15cm}
        \caption{A different time.}
    \end{subfigure}
    \vspace{-.3cm}
    \caption{The trade-off between model complexity and generalization performance across (a) spatial and (b) temporal domain shifts, evaluated on five EfficientNet models (B0-B4) on the representative location of camera trap dataset \cite{lila_datasets}. Fine-tuning is an effective approach for adapting models to new domains across varying model sizes. 
    }
    \label{fig:generalization_vs_capacity}
    \vspace{-0.2in}
\end{figure}

Figure~\ref{fig:generalization_vs_capacity} illustrates the benefits of fine-tuning in addressing domain shifts across different model complexities within the EfficientNet family~\cite{efficientnet}. Models ranging from the lightweight EfficientNet-B0 to the more complex EfficientNet-B4 exhibit substantial gaps in their generalization performance when evaluated on images collected from new locations and different time periods (shown by the blue curve). While larger models inherently possess better domain generalization capabilities, smaller models, though more computationally efficient, tend to perform worse on unseen data.
However, fine-tuning these models with new samples collected from the new locations (Figure~\ref{fig:generalization_vs_capacity}(a)) and different time periods (Figure~\ref{fig:generalization_vs_capacity}(b)) significantly enhances their accuracy, regardless of their computational complexities (shown by the orange curve). \textit{This demonstrates the effectiveness of fine-tuning as a method for adapting models to new domains, bridging the performance gap across varying model sizes.}

\section{Design of \pjn{}}
\label{sec:design}

\begin{figure}[t]
    \centering
    \includegraphics[width=\linewidth]{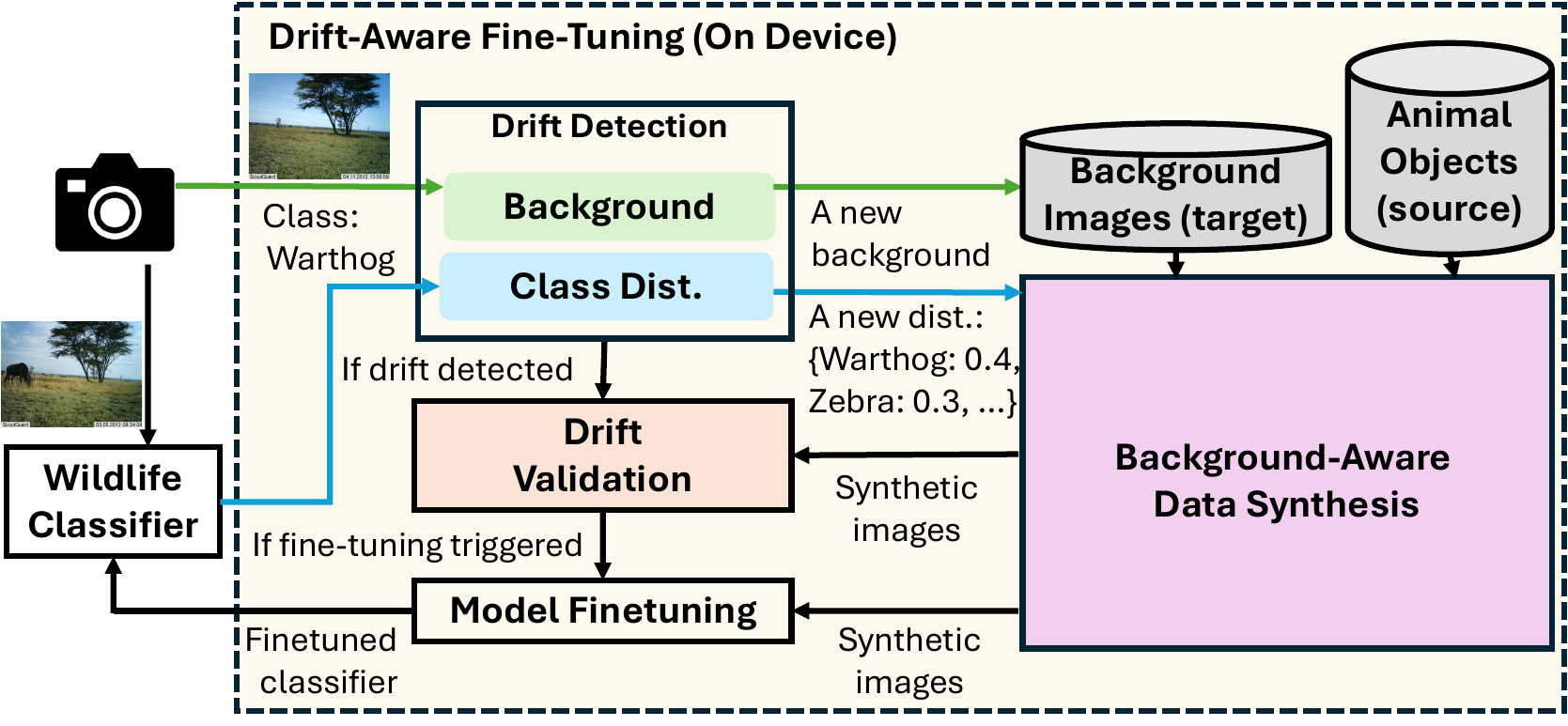}
    \vspace{-.7cm}
    \caption{Overview of \pjn{}. The system runs a lightweight classification model on the IoT device and maintains accuracy under domain shifts through \textit{Drift-Aware Fine-Tuning}, which uses \textit{Background Drift Detection} and \textit{Class Distribution Drift Detection} to identify domain shifts, validates their impact in the \textit{Drift Validation} module, and triggers on-device model adaptation using \textit{Background-Aware Data Synthesis}.}
    \label{fig:overview}
    \vspace{-.3cm}
\end{figure}

\pjn{} is an in-situ fine-tuning system designed to enable camera traps to autonomously and efficiently adapt to domain shifts. We now provide an overview of \pjn{} and then details of the key system components. 

\subsection{Overview of \pjn{}}
\label{sec:overview}

Figure~\ref{fig:overview} illustrates the design of \pjn{}. At its core, \pjn{} runs a lightweight classification model that identifies animals captured by motion-triggered cameras on-device. To maintain accuracy as environmental conditions change, \pjn{} implements a \textit{Drift-Aware Fine-Tuning} pipeline that continuously monitors and responds to domain shifts through three key components:

\mypar{Drift Detection} The system tracks two primary sources of domain shift in wildlife monitoring. \textit{Background Drift Detection} (BDD) identifies environmental changes by analyzing variations in background scenes, while \textit{Class Distribution Drift Detection} (CDD) monitors shifts in the distribution of observed animal species through the model's recent predictions. When either module detects significant changes, it triggers validation to assess the need for model adaptation.

\mypar{Drift Validation} This component serves as an intelligent gate-keeper for model updates and evaluates whether detected shifts actually impact classification performance. It synthesizes domain-specific test data using current backgrounds and class distributions, then measures the model's accuracy to determine if fine-tuning is necessary. This validation step ensures computational resources are only spent on essential model updates.

\mypar{Background-Aware Data Synthesis} When adaptation is needed, this module generates high-fidelity training data by compositing animal objects from the source domain onto background images that reflect current environmental conditions. These synthesized images enable effective model fine-tuning without requiring new labeled animal data from the target domain.

\pjn{} operates autonomously, requiring no manual intervention after deployment. The system maintains a compact repository of source domain animal objects and target domain backgrounds, enabling local synthesis and adaptation. 

\vspace{-0.05in}
\subsection{Background-Aware Data Synthesis}
\label{sec:synthesizer}

\begin{figure}[t]
    \centering
    \includegraphics[width=0.95\linewidth]{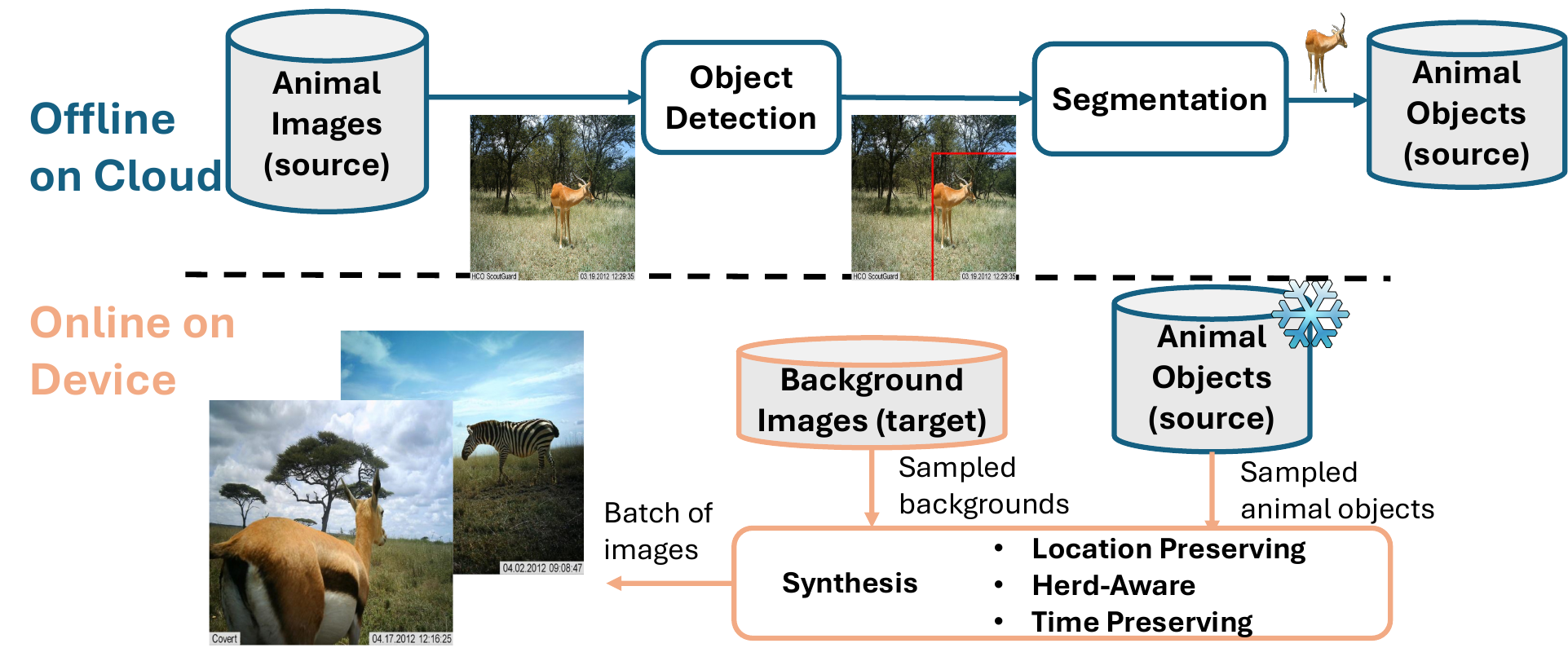}
    \vspace{-.3cm}
    \caption{The background-aware data synthesis. It illustrates two images produced from the Synthesizer. The animal objects repository are frozen on IoT devices.}
    \label{fig:synthesis_pipeline}
    \vspace{-0.2in}
\end{figure}

Background-Aware Data Synthesis (\textit{Synthesizer} in short) addresses a fundamental challenge in adapting on-device classification models: obtaining representative training data from the target domain. While collecting animal images from new environments is impractical and time-consuming, we leverage the fact that background scenes are readily available and capture much of the domain shift in lighting, weather, and seasonal variations. The Synthesizer exploits this opportunity by generating high-fidelity training samples that blend source domain animal characteristics with target domain environmental conditions. Figure~\ref{fig:synthesis_pipeline} illustrates our synthesis pipeline, which combines an offline phase for extracting reusable animal objects with an online phase that creates domain-adapted training data during model fine-tuning.



\mypar{The offline phase} This phase creates a repository of animal objects by extracting them from the training images. 
We utilize MegaDetector V5~\cite{megadetector} to identify the bounding box around each object, then pass the image and the bounding box information to the Segment Anything model~\cite{kirillov2023segment} to extract the object's mask and isolate the object image. 

\mypar{The online phase} 
The online phase generates synthetic training data by intelligently compositing animal objects with target domain backgrounds. Rather than using naive object placement, we leverage domain-specific insights about wildlife behavior and camera trap deployments to improve synthesis quality. Our approach addresses three key challenges in wildlife monitoring: spatial context, social behavior, and temporal patterns. (1) \textit{Location Preserving:} We retain MegaDetector V5 bounding boxes from source-domain images and place animals at the same relative positions on target-domain backgrounds, as their locations are typically tied to terrain features and camera setup. (2) \textit{Herd-Aware:} When source images contain multiple animals (herds), we preserve these group configurations including all bounding boxes. During synthesis, we sample random subsets from these groups to create realistic herd compositions while maintaining natural spatial relationships, effectively modeling social dynamics. (3) \textit{Time Preserving:} We match animal activities with appropriate temporal contexts by pairing diurnal/nocturnal species with corresponding day/night backgrounds. Both the source and target domain data include timestamps, which we leverage to match species with background images captured at corresponding times. These techniques collectively improve fine-tuning accuracy by 4.0\%, as shown in Section~\ref{sec:ablation}.


\subsection{Drift Detection}
Reliable drift detection in images remains an unsolved problem due to the high dimensionality of image data. Traditional methods rely on computationally intensive dimensionality reduction techniques to identify appropriate representations for hypothesis testing~\cite{rabanser2019failing}, rendering them unsuitable for resource-constrained IoT devices.

In contrast to the conventional drift detection methods discussed in Section~\ref{sec:relatedwork}, our approach leverages domain-specific observation that \textit{domain shifts in camera trap data arise from either background (environment) changes or animal class distribution changes}. 
When such deviations are identified, the Drift Detection module signals the need for Drift Validation and, if necessary, Model Fine-Tuning. 

To effectively address these two sources of domain shifts, we designed separate detection modules tailored to each type.

\mypar{Background Drift Detection (BDD)} 
The BDD module monitors changes in background images to identify environmental shifts that may impact model performance. It evaluates whether the current background image significantly differs from those recently collected and, if so, updates the background image repository. 
%

BDD operates on the IoT device by maintaining a sliding window of the last $N$ collected backgrounds ($bg_{pool}$) to compare against each new background ($bg_{cur}$). To ensure temporal relevance, BDD first identifies backgrounds from $bg_{pool}$ that were captured within an hour of $bg_{cur}$ ($bg_{cur \pm 1hr}$). From these candidates, it selects the background with the closest date as the reference ($bg_{ref}$), using capture time as a tiebreaker when needed. The system then employs the Least-Squares Density Difference (LSDD) method~\cite{lsdd,alibi-detect} to compare pixel-level distributions between $bg_{cur}$ and $bg_{ref}$. If the p-test indicates a significant difference (p < 0.05), the new background is added to the device's background repository (Background Images (target) in Figure~\ref{fig:overview}) for use by the Synthesizer.

\begin{figure}[t]
    \centering
    \includegraphics[width=0.95\linewidth]{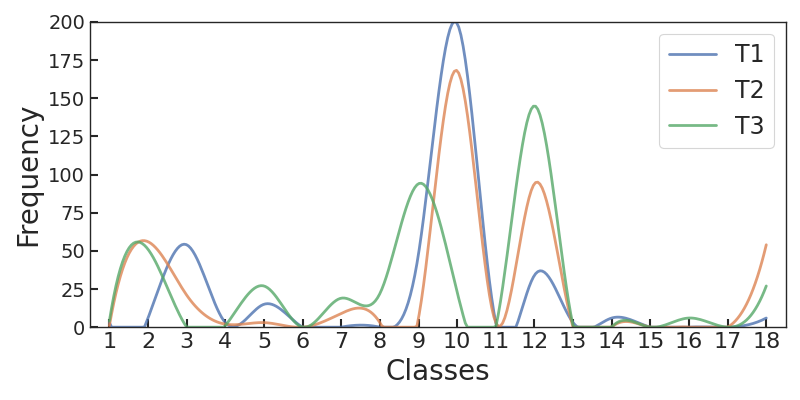}
    \vspace{-.3cm}
    \caption{Class distribution changes across time (location U10 from Serengeti S4 dataset)}
    \label{fig:class_distribution_change}
    \vspace{-0.2in}
\end{figure} 

\mypar{Class Distribution Drift Detection (CDD)}
The CDD module tracks the distribution of predictions on the IoT device and identifies when a drift in class distributions occurs. This module is motivated by the observation that animal class distributions change across locations and over time. 

Figure~\ref{fig:class_distribution_change} shows the class distribution of a test location in the Serengeti S4 dataset~\cite{snapshotserengeti} across three consecutive time windows. The distributions demonstrate progressive divergence over time, reflecting natural changes in animal populations. Specifically, comparing the T1 curve with T2 and T3 reveals increasing divergence as time progresses, with T3 showing a pronounced drift and T2 exhibiting a lighter shift.


CDD operates on the IoT device using a Chi-Squared test, which is particularly suited for detecting shifts in categorical distributions. The module accumulates predictions until it has at least $C$ samples, then compares this recent distribution against the distribution from the last fine-tuning cycle. When the test indicates a significant shift (p < 0.05), CDD triggers Drift Validation to determine if model fine-tuning is necessary.

\subsection{Drift Validation}
\label{sec:drift_val}

While detecting domain shifts is important, not all shifts necessitate model adaptation --- fine-tuning is only warranted when shifts significantly impact classification accuracy. Drift Validation acts as an intelligent gatekeeper, evaluating whether detected shifts will degrade model performance on incoming animal images. Since target domain data is unavailable at validation time, the system leverages the Synthesizer to generate representative test samples that reflect current environmental conditions and class distributions.


However, achieving statistically significant validation requires evaluating the model on many synthesized images, which can be computationally prohibitive for resource-constrained devices. We address this challenge through a \textit{reuse mechanism} that intelligently caches and reuses validation results. By maintaining statistics from previous validations, the system can efficiently estimate model performance on new domains while minimizing redundancy.




The validation process is adaptive to the type of drift detected. For background drift, BDD triggers synthesis of a validation set to assess the current model's accuracy against a threshold ($acc_{ref} - thr$) established during the previous fine-tuning. To minimize computation, we employ a sliding window strategy that synthesizes images only for newly detected backgrounds, reusing inference results from previous validations. When the validation process detects performance degradation and triggers model fine-tuning, the newly fine-tuned model is evaluated on the complete set of synthesized images to establish a new reference accuracy ($acc_{ref}$).

For class distribution drift, CDD leverages pre-computed
per-class accuracies to efficiently estimate overall performance using the latest distribution, avoiding unnecessary image synthesis. The system compares this estimate against the threshold  ($acc_{ref} - thr$) to determine if fine-tuning is needed. Upon initiating a new fine-tuning round, it updates per-class accuracy measurements and establishes a new reference point ($acc_{ref}$) using the current set of synthesized images and the latest class distribution.

Through this approach, Drift Validation efficiently maintains model quality by responding to both environmental and animal distribution changes while minimizing computational overhead.

\subsection{Model Fine-Tuning}
\label{sec:fine-tuning}

Fine-tuning models using images that reflect the current animal class distribution with recent backgrounds may lead to catastrophic forgetting. It is particularly problematic when certain animal classes have not appeared recently, resulting in poor model performance for these classes in the future. 

To prevent catastrophic forgetting while keeping the solution practical for IoT deployments, we implement two simple yet effective techniques that are easy to deploy. 
\textit{First,} we combine recent background images with historical background images to create a more diverse set of synthetic images. These images are stored in Background Images (target) illustrated in Figure~\ref{fig:overview}. This ensures that the model retains knowledge of environmental variations over time, preventing it from becoming overly specialized to the most recent backgrounds. 
\textit{Second,} we adjust the class distribution to prevent the model from over-fitting to recent class occurrences. Mathematically, let $p(c)$ represent the probability of an animal class $c$ appearing in the scene recently, and let $N_{c}$ denote the number of sampled images for class $c$ based on the class distribution. We sample animal classes using a softened distribution defined as $N^{s}_{c} = N_{c} + T$, where $T$ is a tunable temperature parameter that controls the degree of distribution smoothing. By default, we set $T = 20$ throughout our evaluation, which we find to work effectively.

\section{Evaluation}
\label{sec:eval}
This section empirically evaluates the efficacy of \pjn{}. We describe the experiment settings in \S~\ref{sec:settings}, and evaluate Background-Aware Data Synthesis and Drift-Aware Fine-Tuning in \S~\ref{sec:eval_synthesizer}-\ref{sec:eval_finetuning}. We perform end-to-end evaluations and report longitudinal performance of \pjn{} in \S~\ref{sec:end_to_end}-\ref{sec:case_study}, and ablation studies in \S~\ref{sec:ablation}. 

\subsection{Experiment Settings}
\label{sec:settings}
\mypar{Datasets}
We utilize three datasets as shown in Table~\ref{tab:data}: two from the Serengeti Safari Camera Trap network (referred to as D1 and D2)~\cite{snapshotserengeti}, and one from the Enonkishu Camera Trap dataset (referred to as D3)~\cite{snapshotenonkishu}.
These datasets are chosen since they have temporal data over relatively long durations and across several locations captured by trail cameras, hence they exhibit real-world temporal and spatial domain drifts. 
The Camera Trap datasets normally include a lot of species plus an empty class whose images show a scene with no species in it. 
Since some species lack enough samples for both training and testing sets, we selected the 18 most frequent species from the Serengeti dataset and 10 from the Enonkishu dataset for classification and included the empty class to result in 19 and 11 classes respectively. For simplicity, we focused on images containing only one type of animal. Table~\ref{tab:data} summarizes the data statistics. ``Train'' refers to locations whose images are used for training, while ``Test'' represents those for evaluating domain shifts.

\begin{table}[t]
    \centering
    \small 
    \setlength{\tabcolsep}{0.05cm} 
    \begin{tabular}{lcc|cc}
        \hline
        & \multicolumn{2}{c}{\textbf{Train}} & \multicolumn{2}{c}{\textbf{Test}} \\
        \cline{2-3} \cline{4-5}
        \textbf{Dataset} & \textbf{\# Locations} & \textbf{\# Images} & \textbf{\# Locations} & \textbf{\# Images} \\
        \hline
        Serengeti S1 (D1) & 153 & 3692 & 15 & 48063 \\
        Serengeti S4 (D2) & 154 & 6650 & 27 & 82263 \\
        Enonkishu (D3) & 11 & 1043 & 5 & 11003 \\
        \hline
    \end{tabular}
    \caption{Summary of data statistics.}
    \label{tab:data}
    \vspace{-0.4in}
\end{table}

\mypar{\pjn{} Training}
\pjn{} is trained in PyTorch using a two-stage pipeline on the source domain data to ensure high model quality. It employs the EfficientNet-B0 model pre-trained on ImageNet~\cite{deng2009imagenet}. In the first stage, the classification head is trained for 5 epochs with a learning rate (lr) of 1e-3 while keeping the backbone frozen. In the second stage, the entire network is trained for 30 epochs with an lr of 1e-5, using an early stopping criterion of 2 epochs. We use Adam optimizer~\cite{kingma2014adam} and a cross-entropy loss function.

\mypar{In-situ Fine-tuning} For in-situ fine-tuning, we implement several optimizations to enable efficient adaptation on resource-constrained IoT devices.
\textit{First,} to minimize computational overhead, fine-tuning updates only three types of parameters: the fully connected layers, biases, and batch normalization layers. We use an initial learning rate of 1e-4, which is reduced to 1e-5 using a scheduler with a patience of 2 epochs. Early stopping is employed with a threshold of 4 epochs. Both inference and fine-tuning maintain the same input image resolution of $512 \times 512$ as in offline training to ensure domain shifts are not influenced by resolution changes.
\textit{Second,} fine-tuning uses only synthesized images without storing any source domain images, unless noted differently. The background repository is initialized with $N = 80$ images and has maximum capacities of 250, 400, and 140 images for the D1, D2, and D3 datasets, respectively, in order to have balanced classes. As the Background Drift Detection (BDD) module identifies new samples, the oldest background image is replaced to maintain a fixed repository size. 

The fine-tuning process is triggered by two drift detection mechanisms. Class Distribution Drift Detection (CDD) is triggered after every $C = 100$ new predictions. If the dominant class is the empty class, the class distribution is reset. For Drift Validation, we reserve 10\% of objects per class exclusively for validation, while using the remaining data for fine-tuning. The model is fine-tuned only when the Drift Validation module detects a performance drop on synthesized data using the most recent backgrounds and current class distribution. For both BDD and CDD we set the p-value to 5\% and $thr$ for the Drift Validation module is 0\%. We utilized a 2-day window after the BDD triggers a drift and before the Drift Validation module is activated. This ensures that at least 2 days have passed since the last fine-tuning event before Drift Validation is triggered again, effectively controlling the frequency of Drift Validation for background drifts.


\mypar{Platforms} We use a cluster of NVIDIA L40S 48GB GPUs for model training to prepare the initial wildlife classifier. While this offline training uses powerful GPUs, we implement \pjn{}'s in-situ fine-tuning capabilities and evaluate its runtime performance on resource-constrained IoT devices, using EfficientNet-B0 on Raspberry Pi 5, Jetson Orin Nano Super, Xavier NX, and Orin AGX developer kits.



\subsection{Eval. of Background-Aware Synthesis}
\label{sec:eval_synthesizer}

\begin{table}[t]
    \centering
    \small 
    \setlength{\tabcolsep}{0.05cm} 
    \begin{tabular}{c!{\vrule width 1.5pt}c!{\vrule width 1.5pt}c|c!{\vrule width 1.5pt}c|c|c!{\vrule width 1.5pt}c|c|c!{\vrule width 1.5pt}}
        \hline\hline
        & \multicolumn{9}{c}{\textbf{Methods}} \\
        \cline{2-10}
        \textbf{Locs.} & \textbf{Ours} & \textbf{No FT} & \textbf{$\Delta1$} & \textbf{Mix} & \textbf{Cut} & \textbf{$\Delta2$} & \textbf{Obj-St} & \textbf{CC} & \textbf{$\Delta3$} \\
        \hline
        1 & 94.1 & 88.7 & \textbf{+5.4} & 91.3 & 90.5 & \textbf{+2.8} & 91.7 & 92.7 & \textbf{+1.4} \\
        2 & 95.7 & 79.0 & \textbf{+16.7} & 90.5 & 89.5 & \textbf{+5.2} & 92.8 & 94.3 & \textbf{+1.4} \\
        3 & 80.9 & 65.2 & \textbf{+15.7} & 58.2 & 52.6 & \textbf{+22.7} & 66.5 & 69.4 & \textbf{+11.5} \\
        4 & 93.2 & 81.9 & \textbf{+11.3} & 90.2 & 91.3 & \textbf{+1.9} & 90.5 & 92.2 & \textbf{+1.0} \\
        5 & 97.5 & 76.7 & \textbf{+20.8} & 97.9 & 97.5 & \textbf{-0.4} & 97.8 & 98.0 & \textbf{-0.5} \\
        6 & 93.6 & 49.2 & \textbf{+44.4} & 92.4 & 92.3 & \textbf{+1.2} & 93.0 & 93.1 & \textbf{+0.5} \\
        7 & 95.5 & 89.6 & \textbf{+5.9} & 95.4 & 94.7 & \textbf{+0.1} & 95.6 & 95.8 & \textbf{-0.3} \\
        8 & 95.3 & 56.9 & \textbf{+38.4} & 94.2 & 94.0 & \textbf{+1.1} & 94.5 & 94.7 & \textbf{+0.6} \\
        9 & 95.1 & 93.8 & \textbf{+1.3} & 95.4 & 95.2 & \textbf{-0.3} & 95.4 & 95.5 & \textbf{-0.4} \\
        10 & 92.7 & 81.6 & \textbf{+11.1} & 82.9 & 85.1 & \textbf{+7.6} & 88.6 & 90.5 & \textbf{+2.2} \\
        11 & 93.8 & 88.2 & \textbf{+5.6} & 90.8 & 90.9 & \textbf{+2.9} & 92.2 & 93.7 & \textbf{+0.1} \\
        12 & 93.8 & 92.2 & \textbf{+1.6} & 91.6 & 90.1 & \textbf{+2.2} & 90.4 & 93.7 & \textbf{+0.1} \\
        13 & 86.2 & 54.8 & \textbf{+31.4} & 84.0 & 83.6 & \textbf{+2.2} & 82.9 & 84.3 & \textbf{+1.9} \\
        14 & 91.9 & 39.2 & \textbf{+52.7} & 89.4 & 89.3 & \textbf{+2.5} & 89.2 & 90.7 & \textbf{+1.2} \\
        15 & 91.8 & 81.2 & \textbf{+10.6} & 90.6 & 89.5 & \textbf{+1.2} & 91.1 & 91.6 & \textbf{+0.2} \\
        16 & 85.5 & 70.0 & \textbf{+15.5} & 55.3 & 43.5 & \textbf{+30.2} & 71.2 & 72.8 & \textbf{+12.7} \\
        17 & 77.5 & 68.1 & \textbf{+9.4} & 60.5 & 62.5 & \textbf{+15.0} & 69.5 & 71.8 & \textbf{+5.7} \\
        18 & 91.4 & 62.3 & \textbf{+29.1} & 58.0 & 61.3 & \textbf{+30.1} & 80.0 & 79.2 & \textbf{+11.4} \\
        19 & 88.6 & 68.8 & \textbf{+19.8} & 71.8 & 63.8 & \textbf{+16.8} & 83.0 & 84.5 & \textbf{+4.1} \\
        20 & 80.3 & 66.5 & \textbf{+13.8} & 70.0 & 66.4 & \textbf{+10.3} & 73.4 & 76.0 & \textbf{+4.3} \\
        21 & 85.2 & 78.8 & \textbf{+6.4} & 79.9 & 81.1 & \textbf{+4.1} & 80.1 & 82.6 & \textbf{+2.6} \\
        22 & 91.2 & 56.2 & \textbf{+35.0} & 91.6 & 88.7 & \textbf{-0.4} & 90.3 & 91.9 & \textbf{-0.7} \\
        23 & 88.5 & 83.2 & \textbf{+5.3} & 85.6 & 81.6 & \textbf{+2.9} & 85.5 & 87.1 & \textbf{+1.4} \\
        24 & 84.9 & 65.2 & \textbf{+19.7} & 58.3 & 58.6 & \textbf{26.3} & 66.5 & 71.2 & \textbf{+13.7} \\
        25 & 98.5 & 3.8 & \textbf{+94.7} & 98.4 & 98.5 & \textbf{0.0} & 98.8 & 98.9 & \textbf{-0.4} \\
        26 & 88.9 & 79.6 & \textbf{+9.3} & 84.4 & 81.0 & \textbf{+4.5} & 84.4 & 86.1 & \textbf{+2.8} \\
        27 & 91.0 & 69.9 & \textbf{+21.1} & 86.8 & 83.8 & \textbf{+4.2} & 88.6 & 88.3 & \textbf{+2.4} \\
        \hline\hline
        \textbf{All} & 90.5 & 70.0 & \textbf{+20.5} & 82.8 & 81.4 & \textbf{+7.3} & 86.1 & 87.4 & \textbf{+3.0} \\\hline\hline
    \end{tabular}
    \caption{Classification accuracy of our synthesis approach compared to baselines across 27 test locations in the D2 dataset. $\Delta1$ represents the accuracy gain over no fine-tuning (No FT), $\Delta2$ is the gain over the best resource-efficient alternative (MixUp or CutMix), and $\Delta3$ is the gain over the best diffusion-based model (Object-Stitch or ControlCom). The last row shows the average across all locations.}
    \label{tab:eval_synthesizer}
    \vspace{-0.3in}
\end{table}

We start by evaluating the maximum accuracy achievable by each synthesis method when using both source domain and synthesized images to fine-tune the complete EfficientNet-B0 network. The rest of  Sections~\ref{sec:eval_finetuning}-\ref{sec:ablation} examine the more practical IoT scenario where only synthesized images are used and fine-tuning is limited to specific model parameters.

\mypar{Baselines} 
We compare our approach against three classes of techniques that offer different compute-accuracy tradeoffs:

\vspace{.1cm}\noindent(1) \textit{No fine-tuning} (No-FT), which serves as a lower bound baseline. Our goal is to significantly outperform this approach, demonstrating the clear benefits of our synthesized images.

\vspace{.1cm}\noindent(2) \textit{Computationally efficient approaches}, which are efficient enough to feasibly execute on-device on IoT hardware. These include: (a) \textit{CutMix} \cite{yun2019cutmix} (CUT), a data augmentation technique that combines two images by cutting and pasting patches and mixing their labels proportionally. (b) \textit{MixUp} \cite{zhang2017mixup} (MIX), a data augmentation technique that creates new training samples by linearly blending pairs of images and their labels. These approaches are in the same computational ballpark as our techniques. Our objective is to outperform these methods in accuracy while having similar or better computational efficiency.

\vspace{.1cm}\noindent (3) \textit{Diffusion model-based approaches}, which are impractical to execute on-device. These include: (a) \textit{Object-Stitch} (Obj-St)~\cite{objectstitch}, an object compositing method based on conditional diffusion models, specifically designed for blending objects into background images. (b) \textit{ControlCom} (CC)~\cite{controlcom}, a method that synthesizes realistic composite images from foreground and background elements using a diffusion model. Given the significant computational requirements of the diffusion models, we pre-generated all images with both models offline before training. Our goal with \pjn{} is to match or exceed their synthesis quality while providing an considerably more efficient solution that can run directly on IoT devices.


We run this evaluation using the D2 dataset, which has the most test locations. We first randomly selected 250 background images from each test location within the target domain and excluded them from the test set for each location. The selected backgrounds are used to synthesize animal images to fine-tune the classification model. The fine-tuned models are evaluated on images from the target domain. 

\mypar{Classification Accuracy Results}
Table~\ref{tab:eval_synthesizer} compares our synthesis method against baselines. 
We observe the following. 
(1) Compared to no fine-tuning, our synthesis scheme has a massive gain of 20.5\% in accuracy on average (last row), indicating the importance of fine-tuning under domain shifts.
(2) Compared to computationally efficient approaches like CutMix~\cite{yun2019cutmix} and MixUp~\cite{zhang2017mixup}, our method has a significant advantage of 7.3\% on average, with 20-30\% improvements in some locations where the spatial domain shift is large. These generic augmentation methods often strive to augment images in a domain-agnostic way, which can improve model robustness but may not capture the specific characteristics of new deployment environments. 
(3) Compared to diffusion-based methods, our approach achieves a 3\% average accuracy gain (over 10\% in some cases) with much lower computation. This improvement arises because diffusion models often fail to preserve backgrounds and sometimes produce unusable images. Moreover, their slow generation speed prevents on-the-fly synthesis during training, forcing offline image composition. In contrast, our lightweight method supports on-the-fly synthesis for every batch alongside common data augmentation techniques, producing diverse variations (e.g., different herd images) that enhance fine-tuning.

\subsection{Eval. of Drift-Aware Fine-Tuning}
\label{sec:eval_finetuning}

Having established the effectiveness of our synthesis approach, we now evaluate the Drift-Aware Fine-Tuning pipeline that determines when to trigger model updates. This pipeline monitors two types of domain shifts -- background changes and class distribution changes -- and we evaluate how well it balances fine-tuning frequency against model accuracy for each type separately. For fair comparison, all approaches start with identical model initialization.


\mypar{Results under Background Drift}
We first evaluate how well different approaches handle background drift while controlling for class distribution effects. For this comparison, all methods use a uniform animal class distribution and the same initial set of 80 background images. The key difference lies in how they select additional backgrounds over time and decide when to trigger fine-tuning. We compare our approach against the following baselines:

\vspace{.1cm}\noindent(1) \textit{Periodic fine-tuning} (Periodic), where fine-tuning occurs at fixed intervals of N days. The fine-tuning process utilizes randomly sampled background images. To ensure a fair comparison, the number of randomly sampled backgrounds is matched to the number of backgrounds detected by BDD. A larger N results in less fine-tuning but risk higher accuracy drops under background drifts.

\vspace{.1cm}\noindent(2) \textit{Periodic fine-tuning with BDD} (Periodic+BDD): This baseline performs fine-tuning at fixed intervals but uses backgrounds selected by our Background Drift Detection module rather than random sampling. By comparing it against Periodic, we can isolate the value of BDD's background selection strategy. Comparing it against our full approach reveals the additional benefits of using Drift Validation to trigger fine-tuning only when necessary.

\begin{figure}[t]
    \centering
    \begin{subfigure}[t]{0.23\textwidth}
        \centering
        \includegraphics[height=1.2in]{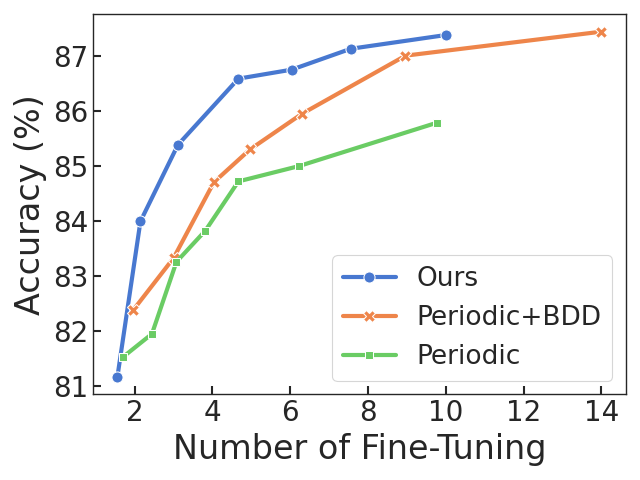}
        \vspace{-.4cm}
        \caption{Background Drifts.}
        \label{fig:bdd}
    \end{subfigure}
        \centering
    \begin{subfigure}[t]{0.23\textwidth}
        \centering
        \includegraphics[height=1.2in]{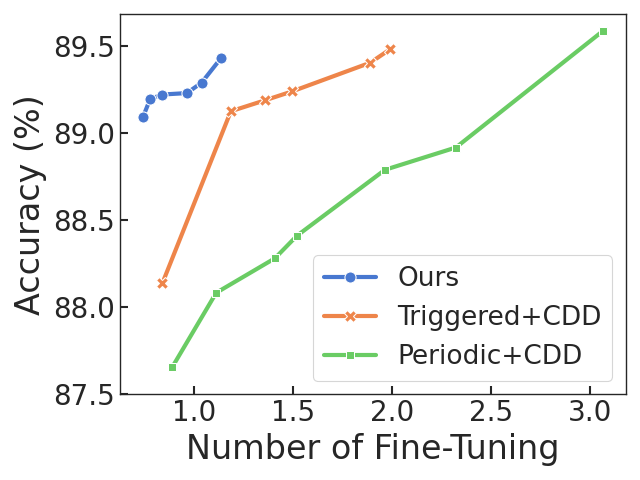}
        \vspace{-.4cm}
        \caption{Class distribution drifts.}
        \label{fig:cdd}
    \end{subfigure}
    \vspace{-0.3cm}
    \caption{Tradeoff between model accuracy and fine-tuning frequency under background drift (a) and class distribution drift (b). Drift-aware fine-tuning triggers the least fine-tuning while achieving higher accuracy than counterparts. Results are averaged over 27 locations in the D2 dataset.}
    \vspace{-0.2in}
\end{figure}


Figure~\ref{fig:bdd} illustrates the accuracy-efficiency tradeoffs achieved by different approaches. For the Periodic Fine-Tuning baselines, we vary the intervals between fine-tuning events, setting $N = {2, 4, 6, 8, 10, 14, 21}$ days to achieve different trade-off points. 
In contrast, our approach adjusts the thresholds for acceptable accuracy degradation within the Drift Validation module, utilizing thresholds $thr$ from $-6\%$ to $0\%$ in $1\%$ increments to explore trade-off scenarios. 


Overall, our approach achieves the Pareto frontier of fine-tuning frequency and model accuracy under background drift.    
By comparing \textit{Periodic + BDD} (orange curve) against \textit{Periodic} (green curve), we observe that BDD  outperforms random sampling (up to 2\%), particularly as the number of fine-tuning rounds increases (smaller N). This highlights the effectiveness of BDD in detecting background drift. 
Comparing our approach with \textit{Periodic + BDD}, our method achieves higher accuracy (up to 1.5\%) for the same number of fine-tuning rounds. To reach a specific accuracy, our approach requires fewer fine-tuning rounds (over 4 rounds) highlighting the effectiveness of the Drift Validation module. This reduction in fine-tuning frequency is especially useful in unattended settings, where minimizing computational overhead and preserving battery life are paramount.

\mypar{Results under Class Distribution Drift}
To ensure that the fine-tuning and accuracy evaluation are not influenced by background drifts, we pre-select 250 background images for image synthesis across all approaches. We compare our fine-tuning strategy (\textit{ours}) against the following baselines.   

\vspace{.1cm}\noindent(1) \textit{Periodic fine-tuning with CDD} (Periodic+CDD), where fine-tuning happens periodically after N days. It uses the smoothed class distribution in Section~\ref{sec:fine-tuning} in synthesizing training images. 
 
\vspace{.1cm}\noindent(2) \textit{Fine-tuning triggered with CDD} (Triggered+CDD), where fine-tuning happens once a class distribution drift is detected. Comparing this baseline to ours highlights the effectiveness of the Drift Validation module. 

Figure~\ref{fig:cdd} reports the trade-off between fine-tuning frequency and model accuracy for our proposed approach and the baselines under class distribution drift. 
For the Periodic Fine-Tuning baselines, the intervals between fine-tuning events are varied, with $D = {4, 7, 10, 14, 18, 21, 28}$ days, to explore different trade-off points. For Triggered+CDD, the p-value in the p-test is varied, with $p = {0.5\%, 1\%, 2\%, 3\%, 4\%, 5\%}$.
In contrast, our approach adjusts the thresholds for acceptable accuracy degradation within the Drift Validation module, utilizing thresholds $thr$ from $-6\%$ to $0\%$ in $1\%$ increments as before to explore various trade-off scenarios. 



Overall, our approach achieves the Pareto frontier of fine-tuning frequency and model accuracy under class distribution drifts.  
Comparing \textit{Triggered + CDD} and \textit{Periodic + CDD}, we observe that \textit{Triggered + CDD} achieves the same accuracy with fewer fine-tuning rounds (up to 1.5 fewer on average), demonstrating the CDD module's effectiveness in identifying necessary class distribution shifts and triggering fine-tuning. Furthermore, comparing \textit{Ours} and \textit{Triggered + CDD} reveals that the Drift Validation module enhances the performance when paired with CDD, achieving higher accuracy (up to 1\%) with the same number of fine-tuning rounds.

\subsection{End-to-End Performance}
\label{sec:end_to_end}

This section reports the end-to-end evaluation of \pjn{}'s performance in terms of accuracy and runtime performance.

\mypar{Results on Accuracy} 
For model accuracy, we compare the following approaches: (1) \textit{Pseudolabel:} This baseline assumes that a few target domain animal images are collected for model fine-tuning. As ground truth labels are unavailable, it uses the IoT model (EfficientNet-B0), trained on the source domain, to generate pseudo-labels for fine-tuning, following a Test-Time Adaptation (TTA) approach. (2) \textit{GTlabel:} This approach assumes that the target domain animal images have ground truth labels. While impractical, it gives us an upper-bound for the Pseudolabel method. (3) \textit{\pjn{}:Syn:} This approach removes drift-aware fine-tuning for domain adaptation. We start by initializing our background repository with the first 80 samples and perform a single round of fine-tuning. This baseline addresses the spatial domain shift only. (4) \textit{\pjn{}:Syn+BDD:} This approach removes only the CDD module. Like \pjn{}-Syn, the background repository is initialized with 80 images, but new backgrounds are added by BDD. Fine-tuning is triggered by the Drift Validation module in response to detected background drift, with no consideration of class distribution shifts, and (5) \textit{\pjn{}:All:} This is the complete \pjn{}.







For fair comparison, we control how much data each method uses and when fine-tuning occurs. The \textit{Pseudolabel} and \textit{GTlabel} approaches sample the same number of images as BDD collects backgrounds in \pjn{}, and all methods use \pjn{}'s fine-tuning triggering mechanism to determine when to update their models. While using \pjn{}'s timing for fine-tuning gives an advantage to the baselines, it ensures a controlled evaluation where all methods perform the same number of updates at the same points in time.



\begin{table}[t]
    \centering
    \small 
    \setlength{\tabcolsep}{0.2cm} 
    \begin{tabular}{lccc}
        \hline
        & \multicolumn{3}{c}{\textbf{Datasets}} \\
        \cline{2-4}
        \textbf{Methods} & \textbf{D1} & \textbf{D2} & \textbf{D3} \\
        \hline
        Pseudolabel & 55.4 & 58.3 & 39.3 \\
        GTlabel & 66.0 & 71.3 & 53.1 \\ \hline 
        \pjn{}:Syn & 79.8 & 74.7 & 66.0 \\
        \pjn{}:Syn+BDD & 91.1 & 87.1 & 75.8 \\
        \pjn{}:All & 91.6 & 88.9 & 76.2 \\

        \hline
    \end{tabular}
    \caption{Mean accuracy across test locations for end-to-end approaches. 
    }
    \label{tab:end_to_end}
    \vspace{-.3in}
\end{table}

The results in Table~\ref{tab:end_to_end} reveal several important insights about the accuracy of different approaches. 

\textit{Comparison with Pseduolabel:} The Pseduolabel approach performs poorly in all datasets with accuracies 55.4\%, 58.3\% and 39.3\% for D1, D2 and D3, respectively. This underperformance is primarily due to the inaccuracy of the pseudo-labels as the starting model is only trained on the source domain, highlighting the importance of proactive fine-tuning rather than relying on reactive approaches.

\textit{Comparison with GTlabel:} Despite utilizing ground truth labels in the GTlabel approach, performance remains suboptimal: accuracy reaches only 66.0\%, 71.3\%, and 53.1\% for D1, D2, and D3, respectively. 
This is due to the infrequent and inconsistent appearance of animals. Different species emerge at different times, posing a challenge in collecting sufficient labeled data that ensuring each animal class is adequately represented for effective fine-tuning.

Compared to both the above methods, \pjn{} (last row) is significantly better with average accuracies of 91.6\%, 88.9\%, and 76.2\% for D1, D2, and D3 respectively. 


\vspace{.1cm}\noindent \textit{\pjn{}'s Performance Breakdown:} The last three rows of Table~\ref{tab:end_to_end} demonstrates that each stage of \pjn{} contributes to performance improvement. The background-aware synthesis approach (Syn) alone outperforms the performance of GTlabel and achieves 79.8\%, 74.7\%, and 66.0\% accuracy for D1, D2, and D3 respectively. Adding Background Drift Detection (BDD) with the Drift Validation Module, which enables model fine-tuning over time, results in significant gains, improving accuracy by 11.3\%, 12.4\%, and 9.8\% for D1, D2, and D3 respectively. The inclusion of Class Distribution Drift Detection (CDD) completes our full \pjn{} approach and adds about 0.4-1.8\% accuracy improvement.

\begin{table}[t]
\centering
\small
\setlength{\tabcolsep}{0.1cm} 
\begin{tabular}{lcccc}
\hline
\textbf{} & \textbf{RPi-5} & \textbf{O.S. Nano} & \textbf{X. NX} & \textbf{O. AGX} \\
\hline
\multicolumn{5}{c}{\textit{Synthesis Methods}} \\
\hline
\textbf{\pjn{}:Syn} (CPU) & 61.6 ms & 10.2 ms & 149.7 ms & 60.2 ms \\
CutMix:Syn (CPU) & 69.9 ms & 6.9 ms & 445.4 ms & 36.7 ms \\
MixUp:Syn (CPU) & 100.2 ms & 9.2 ms & 446.5 ms & 43.2 ms \\
Object-Stitch:Syn (GPU) & - & - & - & 319.1 s \\
Control-Com:Syn (GPU) & - & - & - & 613.2 s \\
\hline
\hline
\multicolumn{5}{c}{\textit{Fine-Tuning}} \\
\hline
\textbf{\pjn{}:Fine-Tuning} & 67.65 s & 1.98 s & 3.33 s & 1.75 s \\
\hline
\end{tabular}
\caption{Latency for synthesis and Fine-Tuning (one iteration) on a batch of 32 images is compared across \pjn{} and alternative approaches on four devices: Raspberry Pi 5 (8GB), Jetson Orin Nano Super (8GB), Jetson Xavier NX (8GB), and Jetson Orin AGX (32GB). CPU-based approaches for image synthesis use \#workers = 4.}
\label{tab:performance_comparison}
\vspace{-.3in}
\end{table}

\mypar{Runtime Performance Results} Table~\ref{tab:performance_comparison} reports the runtime performance of \pjn{} on four platforms representing a range of IoT deployment scenarios. We evaluated on the Raspberry Pi 5 for ultra-low-power deployments, the Jetson Orin Nano Super as a representative of current-generation edge AI devices that we expect to be widely deployed in future IoT systems, the Jetson Xavier NX for mid-range applications, and the Jetson Orin AGX for high-performance edge computing. To fine-tune the model on the Raspberry Pi 5, Jetson Orin Nano, and Jetson Xavier NX with an effective batch size of 32 without OOM, we adopted a gradient accumulation technique that uses a batch size of 4 with 8 accumulation steps to update the weights. The synthesis latency is the time spent on the image synthesis and does not include the latency introduced by the dataloader.

Overall, the results show that \pjn{} has very low computational overhead and is highly practical for real-world use on resource-constrained platforms while allowing our wildlife classification system to continuously adapt to changing environments.
The performance on the Jetson Orin Nano Super is particularly promising, with synthesis taking only 10.2 milliseconds and fine-tuning requiring just 1.98 seconds per iteration, demonstrating excellent practicality for real-world IoT deployments. We see that \pjn{} is extremely efficient in generating synthetic images, introducing minimal overhead for model fine-tuning. For example, on the Raspberry Pi 5, \pjn{} spends 61.6 milliseconds for synthesizing a batch of 32 images, which is approximately three orders of magnitude faster than the 67.65 seconds required for training a single batch with EfficientNetB0. Similarly, on the Jetson Orin Nano Super, synthesis takes only 10.2 milliseconds, roughly three orders of magnitude less than the 1.98 seconds for fine-tuning. This high efficiency is consistent on other tested devices as well.
Although data augmentation methods like CutMix and MixUp are comparably efficient for synthesis, they result in worse model performance due to the lower quality of the synthesized data, as discussed in \S~\ref{sec:eval_synthesizer}.

In contrast, diffusion-based image composition techniques, despite utilizing GPUs, are exceedingly slow and cannot be executed on embedded devices (Pi5, Nano, NX). For instance, Control-Com requires 613.2 seconds to synthesize a batch of 32 images even on the powerful Orin AGX, making it thousands of times slower than \pjn{} on any of our tested devices. Control-Com's synthesis time on the Orin AGX is 9$\times$ slower than fine-tuning on Pi 5, 310$\times$ slower than fine-tuning on Orin Nano Super, 184$\times$ slower than fine-tuning on Xavier NX, and 350$\times$ slower than fine-tuning on Orin AGX itself. This extreme latency makes diffusion-based methods impractical for real-time or on-device adaptation in IoT settings.

\mypar{Results on Memory Requirements} 
Memory usage varies due to hardware constraints and batch size adaptation. The Orin AGX uses a batch size of 32 (15.1 GB GPU, 5.5 GB CPU), while other devices use a batch size of 4 with gradient accumulation for effective batch size of 32: Raspberry Pi 5 requires 4 GB CPU memory, Orin Nano Super uses 2 GB GPU and 2.5 GB CPU, and Xavier NX uses 2 GB GPU and 3.5 GB CPU memory. The difference in CPU memory usage between the Nano Super and Xavier NX reflects their different CPU architectures and cache efficiencies, with the Nano Super benefiting from a newer CPU generation.

\subsection{Case Study: Longitudinal Performance}
\label{sec:case_study}
To further illustrate how \pjn{} responds to domain shifts, we present the longitudinal system performance of \pjn{} on a real-world camera trap trace from the D2 dataset. 

\mypar{Results on Accuracy}
We compare \pjn{} against the following two alternative methods:

\vspace{.1cm}\noindent(1) \textit{Oracle}: This method assumes that the wildlife classifier has been fully trained using the target domain data, which is hypothetically known beforehand. For this, we randomly sampled 10\% of the data from the entire trace for model training, reserving the remaining 90\% for testing. To mitigate the limited size of the training dataset, we supplemented the training data with synthesized samples generated by our Synthesizer. Although this approach is infeasible in practice due to the unavailability of target domain data a priori, it provides a reference for model performance on a noisy, real-world animal classification dataset. Our goal is to achieve comparable accuracy as this approach while offering a practical solution to overcome domain shifts.

\vspace{.1cm}\noindent(2) \textit{\pjn{}:Syn}: This method is the same as \pjn{}:Syn (Table~\ref{tab:end_to_end}). 



\begin{figure}[t]
    \centering
    \includegraphics[width=0.97\linewidth]{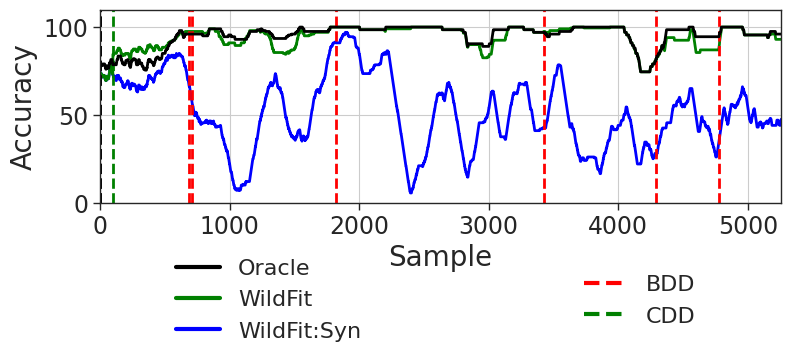}
    \vspace{-.3cm}
    \caption{Longitudinal system performance on the location "R09" of D2 dataset. A total of 7 fine-tuning events are triggered as shown by the vertical dotted lines (BDD triggers are in red, and CDD triggers in green). Two BDD events occur close together, forming the second thick red dotted line.}
    \label{fig:longitudinal}
    \vspace{-0.2in}
\end{figure}

Figure~\ref{fig:longitudinal} shows how  accuracy changes over time for a test location from the D2 dataset. Each point on the accuracy curves represents the performance of a sliding window comprising 200 reserved test samples. Dotted lines on the plot indicate fine-tuning events triggered by the Background Drift Detection (BDD) and Class Distribution Drift Detection (CDD) modules, in conjunction with the Drift Validation module. All approaches were evaluated on the reserved 90\% of target domain data, ensuring a fair and statistically significant accuracy comparison.

We observe that \pjn{} closely tracks the performance of \textit{Oracle}, demonstrating the effectiveness of \pjn{}'s in-situ fine-tuning in addressing domain shifts. The significant performance gap between \textit{Oracle} and \textit{\pjn{}-Syn} further supports this point, as \textit{\pjn{}-Syn} does not handle temporal domain shifts. Notably, \textit{\pjn{}-Syn} initially aligns with the performance of \pjn{} prior to the first model fine-tuning event triggered by a class distribution shift (indicated by the first green dotted line from the CDD module). After this point, \textit{\pjn{}-Syn} exhibits substantially worse performance over time, highlighting the importance of continuous adaptation in dealing with real-world domain shifts.

\mypar{Fine-Tuning Time} 
In terms of end-to-end fine-tuning time, a total of 7 fine-tuning events are triggered as shown by the vertical dotted lines in Figure~\ref{fig:longitudinal}. Fine-tuning times range from 2.4 to 5.3 minutes (average: 3.1 minutes) on the Jetson Orin Nano Super, 3.5 to 7.8 minutes (average: 5.4 minutes) on the Orin AGX,  and 11.3 to 21.6 minutes (average: 14.3 minutes) on the Jetson Xavier NX. Since the underlying camera trap system is motion-triggered and animals arrive on average every 22.5 minutes at this location, the fine-tuning duration remains well within reasonable bounds relative to the natural arrival rate of wildlife events.

\mypar{Latency and Energy Under Different Power Modes} 
In this subsection, we evaluate the latency and energy consumption of the major components of the \pjn{} system under different power modes of the Jetson Orin Nano Super (8 GB). Table~\ref{tab:power_mode} provides a detailed breakdown for the 15W, 25W, and MAXN (Super) configurations. The MAXN Super mode represents an uncapped power setting that enables the maximum number of active cores and highest clock frequencies across the CPU, GPU, DLA, PVA, and SoC components. As expected, higher power modes lead to reduced latency, particularly for model fine-tuning, with only a modest increase in energy consumption. Compared to the 15W mode, the 25W and MAXN Super modes achieve 1.4$\times$ and 1.5$\times$ lower latency, respectively, while incurring just 3.5$\%$ and 5.9$\%$ increases in energy usage. This demonstrates excellent energy efficiency that allows deployments to optimize for either battery life (15W) or performance (MAXN) with minimal energy penalties. The results also highlight the efficiency of the CDD and BDD modules, which incur minimal overhead. Additionally, the validation module demonstrates both effectiveness and efficiency, leveraging a reuse mechanism and precomputed per-class accuracies (\S~\ref{sec:drift_val}) to avoid unnecessary fine-tuning. Most importantly, the validation module consumes up to 200$\times$ less energy than a single fine-tuning round, making it highly cost-effective for filtering out unnecessary adaptations and preserving battery life in remote deployments.

\begin{table}[t]
    \centering
    \small
    \setlength{\tabcolsep}{0.04cm}
    \begin{tabular}{lc|cc|cc|c}
        \toprule
        \multirow{2}{*}{\textbf{Components}} & \multicolumn{2}{c}{\textbf{15W}} & \multicolumn{2}{c}{\textbf{25W}} & \multicolumn{2}{c}{\textbf{MAXN}} \\
        \cmidrule(lr){2-3} \cmidrule(lr){4-5} \cmidrule(lr){6-7}
        & \textbf{Latency} & \textbf{Energy} & \textbf{Latency} & \textbf{Energy} & \textbf{Latency} & \textbf{Energy} \\
        \hline
        CDD  & 1.8 ms & 15.8 mJ & 1.8 ms & 22.6 mJ & 1.5 ms & 21.5 mJ \\
        BDD  & 68.0 ms & 467.2 mJ & 70.2 ms & 498.3 mJ & 63.6 ms & 548.6 mJ \\
        Inference  & 62.8 ms & 440.3 mJ & 61.1 ms & 455.7 mJ & 54.9 ms & 496.6 mJ \\
        Validation  & 1.7 s & 13.8 J & 1.4 s & 13.2 J & 1.3 s & 15.1 J \\
        Fine-Tuning  & 274.9 s & 2634.1 J & 191.6 s & 2730.4 J & 182.3 s & 2789.2 J \\
        \bottomrule
    \end{tabular}
    \caption{Average latency and energy consumption of each component in the \pjn{} pipeline under different power modes, implemented and evaluated on the Jetson Orin Nano (8 GB) for location "R09" of D2 dataset.}
    \label{tab:power_mode}
    \vspace{-0.3in}
\end{table}

\begin{figure}[t]
    \centering
    \includegraphics[width=0.8\linewidth]{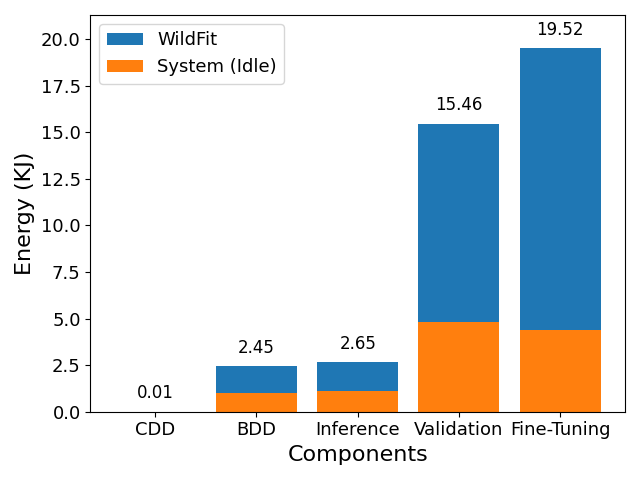}
    \vspace{-.5cm}
    \caption{Total energy consumption of each component in the end-to-end implementation of \pjn{} on the Jetson Orin Nano Super (8 GB) for location R09 of the D2 dataset, collected over a 37-day period. Orange bars indicate the energy consumed by the system while powered on but idle (i.e., not performing any computation).}
    \label{fig:total_energy}
    \vspace{-0.1in}
\end{figure}

\mypar{Results on Total Energy Consumption} 
We now report the total energy consumption of the major components of the \pjn{} system, implemented on the Jetson Orin Nano Super (MAXN mode), for location R09 of the D2 dataset, collected via a camera trap over a 37-day period. Figure~\ref{fig:total_energy} illustrates the energy breakdown, where approximately one-third to one-fourth of each bar represents system idle energy consumption, with the remainder attributed to active computation. As observed earlier, both the CDD and BDD modules remain highly lightweight in terms of energy usage. Notably, BDD is triggered more frequently due to background dynamics. The total energy consumption for 7 rounds of model fine-tuning was 19.52 kJ. In contrast, the validation module—while incurring energy equivalent to approximately 5 fine-tuning rounds—was triggered nearly 1000 times, successfully filtering out many unnecessary fine-tuning attempts. This highlights its effectiveness in verifying model robustness against minor temporal, spatial, and class distribution shifts, thereby preventing excessive energy expenditure.


\mypar{Results on Battery Requirements} We evaluated energy consumption by running \pjn{} on the Jetson Orin Nano Super with a 37-day image stream from the location ``R09'' in the D2 dataset. The system consumed only 11.2 Wh total, demonstrating exceptional energy efficiency. This low power requirement means the system can operate on just three standard AA batteries, or run for approximately six months on a compact 10,000 mAh power bank, making it highly practical for remote wildlife monitoring deployments.

\subsection{Ablation Studies}
\label{sec:ablation}


%

\begin{table}[t]
    \centering
    \small 
    \setlength{\tabcolsep}{0.2cm} 
    \begin{tabular}{lccccc}
        \hline
        & \multicolumn{5}{c}{\textbf{\#Animal Objects Per Class}} \\
        \cline{2-6}
        \textbf{Methods} & 100 & 150 & 250 & 300 & 350 \\
        \hline
        No-FT (Acc.) & 70.0 & 70.0 & 70.0 & 70.0 & 70.0 \\\hline
        \pjn{} (Acc.) & 83.9 & 84.1 & 84.4 & 85.6 & 86.9 \\\hline
        Accuracy gain & +13.9 & +14.1 & +14.4 & +15.6 & +16.9 \\
        \hline
    \end{tabular}
    \caption{Effect of the number of animal objects per class stored on a device on model accuracy (averaged over 27 test locations of D2 dataset).}
    \label{tab:ablation_tr_size}
    \vspace{-0.2in}
\end{table}

\textbf{Impact of the Number of Animal Objects.} 
Background-aware data synthesis requires a repository of animal objects stored on device for data synthesis. More animal objects per class introduces higher storage overhead but improves synthesized image diversity and thus improves fine-tuning performance. 
Table~\ref{tab:ablation_tr_size} evaluates the effect of the number of animal objects per class stored on the device. The model is trained once using only synthesized data (generated from 250 backgrounds), without incorporating the BDD, CDD, or Drift Validation modules. The results highlight \pjn{}'s robustness to reduced training data; as the number of training instances per class decreases from 350 to 100, \pjn{} shows only a 3.0\% drop in accuracy.


\begin{table}[t]
    \centering
    \small 
    \setlength{\tabcolsep}{0.09cm} 
    \begin{tabular}{lcccc}
        \hline
        \textbf{Trainable Parameters} & \textbf{Accuracy} & \multicolumn{3}{c}{\textbf{Latency (s)}} \\
        \cline{3-5}
        & & \textbf{Nano} & \textbf{NX} & \textbf{AGX} \\
        \hline
        FC & 80.2 & 0.79 & 1.27 & 0.61 \\
        FC + Bias & 85.3 & 1.94 & 3.31 & 1.70 \\
        \textbf{FC + Bias + BatchNorm (ours)} & 86.9 & 1.98 & 3.33 & 1.75 \\ \hline 
        Whole & 89.5 & 2.20 & 4.01 & 1.99 \\
        \hline
    \end{tabular}
    \caption{Accuracy and latency of fine-tuning different parameters of EfficientNet-B0 (averaged over 27 test locations of the D2 dataset). Latency represents the time spent on fine-tuning for one iteration using a batch size of 32 on 512×512 images.}
    \label{tab:layer_accuracy}
    \vspace{-0.3in}
\end{table}

\mypar{Impact of Parameters to Fine-Tune} Table~\ref{tab:layer_accuracy} summarizes the impact of parameters to fine-tune on accuracy and one-iteration latency, averaged across 27 test locations in the D2 dataset. Fine-tuning the Fully Connected (FC) layers, biases, and Batch Normalization (BN) layers achieves an effective balance between accuracy and latency for on-device fine-tuning. In particular, updating the BN layers is inspired by techniques from the Test-Time Adaptation (TTA) literature.
The training latency for updating only the FC layers and biases is comparable to our approach, but including Batch Normalization layers adds a 1.6\% accuracy improvement. While training only the FC layers offers faster performance, it results in a significant 6.7\% accuracy drop compared to our approach. 


\begin{table}[t]
    \centering
    \small 
    \setlength{\tabcolsep}{0.04cm} 
    \begin{tabular}{l|c|c}
        \hline
        \textbf{Techniques} & Acc.  & Gain \\
        \hline
        Random Placement & 82.9 & \\
        Loc. Preserving & 84.0 & +1.1\\
        Loc. Preserving + Herd & 85.7 & +2.8 \\
        \textbf{Loc. Preserving + Herd + Time Preserving (Ours)} & 86.9 & +4.0 \\
        \hline
    \end{tabular}
    \caption{Effect of each synthesis technique used in \pjn{} on accuracy (\%, averaged over 27 test locations of D2 dataset).}
    \label{tab:synthesis_techn}
    \vspace{-0.3in}
\end{table}

\mypar{Impact of Synthesis Techniques} Table~\ref{tab:synthesis_techn} highlights how each technique in background-aware data synthesis improves fine-tuning performance. results are averaged across 27 test locations in the D2 dataset.
Random placement is a baseline where a sampled animal object is randomly placed on a background image. Location-preserving technique provides an 1.1\% accuracy boost. Herd-awareness further improves accuracy by 1.7\%, and time-preserving synthesis adds an additional 1.2\%.

\section{Related Work}
\label{sec:relatedwork}

\textbf{Domain Adaptation.} Domain adaptation (DA) focuses on adapting a model trained on the source domain to perform well on a different but related target domain. 
Typically, DA methods assume the availability of labeled or unlabeled target data for model adaptation, making them a poor fit for wildlife monitoring applications where data collection is labor-intensive. 
Some DA approaches, known as \textit{zero-shot DA}, attempt to generalize to unseen target domains by leveraging \textit{auxiliary information} about the target domain even if they don't have direct access to target domain data during training. For example, ZDDA~\cite{peng2018zero} and CoCoGAN~\cite{wang2019conditional} learn from the task-irrelevant dual-domain pairs, while Poda~\cite{fahes2023poda} uses natural language descriptions of the target domain. These approaches cannot be applied, as they rely on auxiliary information that is unavailable for wildlife animal classification tasks. Recently proposed methods, including ZoDi~\cite{zodi}, Min-Max~\cite{daynight1}, and CIConv~\cite{daynight2}, target zero-shot adaptation for day-night or weather-induced shifts. However, these methods assume consistent scene layouts, which are unsuitable for wildlife scenarios and incur heavy computational overhead, making them impractical for resource-constrained deployments.

\mypar{Context-Awareness} Related to our work is the idea of context-aware inference~\cite{rastikerdar2024cactus, palleon, FAST} which aims to improve model performance by dynamically switching between specialized models trained for different operational contexts. Our work takes a fundamentally different approach --- instead of relying on target domain data collection, we enable proactive adaptation through efficient data synthesis that anticipates domain shifts before they occur. This distinction is particularly important where target domain data is scarce and domain shifts are frequent (e.g. in wildlife monitoring). While context-aware approaches require substantial data collection, our synthesis-based approach can continuously adapt even when target domain data is limited or unavailable.

\mypar{Domain Generalization}  
Domain generalization (DG) aims to create models that are inherently robust to domain shifts by ensuring good performance across various source domains. Unlike DA, DG assumes no access to target domain data during training.
DG methods can be categorized into domain alignment~\cite{muandet2013domain}, meta-learning~\cite{li2018learning}, ensemble learning~\cite{zhou2012ensemble}, and foundation models~\cite{bommasani2021opportunities}. 
However, achieving strong generalization often requires computationally complex models, as seen in recent foundation models~\cite{bommasani2021opportunities}. In Section~\ref{sec:background}, we demonstrated that larger models can still benefit from fine-tuning in adapting to domain shifts.

\mypar{Data Augmentation} 
Data augmentation approaches can generally improve the robustness of ML models. The existing literature roughly falls into the following groups: hand-engineered transformations~\cite{shi2020towards}, adversarial attacks~\cite{volpi2018generalizing}, learned augmentation models~\cite{huang2017arbitrary, choi2019self}, feature-level augmentation~\cite{zhang2017mixup, yun2019cutmix}, and more recently generative models such as diffusion models~\cite{trabucco2023effective}.
The most relevant for us are feature-level augmentation techniques such as CutMix~\cite{yun2019cutmix} and MixUp~\cite{zhang2017mixup} that generate synthetic images by blending objects of interest with background images. Despite their efficiency, we show that these generic techniques do not produce high-quality synthetic animal images, resulting in limited, if any, performance improvement. 
Diffusion models, on the other hand, offer the potential for high quality and realistic image synthesis that can better match target scenes~\cite{controlcom, objectstitch, libcom, AnyDoor}. 
However, these incur immense computational cost making them infeasible for rapid, in-situ adaptation to changing scenes. 


\mypar{Drift Detection}  
Drift detection in ML traditionally focuses on identifying changes in data distribution that can impact model performance over time. The literature includes various approaches, such as statistical hypothesis testing, distance-based metrics, and ensemble methods~\cite{alibi-detect, lipton2018detecting}. 
\pjn{} leverages statistical methods such as Least-Squares Density Difference (LSDD), which are frequently employed to detect shifts in data streams~\cite{rabanser2019failing}. These methods are chosen for their computational efficiency, making them suitable for execution on IoT devices. 
What sets \pjn{} apart is its complete pipeline that combines drift detection, validation, and synthetic data fine-tuning to autonomously address domain shifts.

\section{Discussion}
\label{sec:discussion}
\label{sec:conclusion}

\mypar{Test-Time Adaptation (TTA)} TTA~\cite{tta1,SHOT,tta3,tent} enables a model to adjust to distribution shifts during inference by adapting a pre-trained source model to unlabeled target data before making predictions. The key challenge in TTA is achieving robust adaptation with limited or no labeled target samples, a significant problem in applications like wildlife monitoring where data collection is expensive and labels are scarce. Existing TTA methods address this challenge through several strategies, each with distinct trade-offs. Entropy minimization approaches~\cite{tent} encourage confident predictions by reducing output entropy on test samples, but become unstable when target data is limited. Normalization-based methods~\cite{AdaBN} realign feature distributions by updating batch normalization layers, offering computational efficiency but only correcting shallow, feature-level shifts that fail under strong domain changes. Pseudo-labeling approaches~\cite{SHOT} generate labels for target samples to guide self-training, yet reinforce errors when initial predictions are uncertain. We demonstrate this in our evaluation, where the Pseudolabel baseline (a TTA approach) performs poorly. More recent continual TTA methods~\cite{CoTTA} adapt sequentially in non-stationary environments, but require maintaining multiple models (e.g., student and teacher) and remain prone to catastrophic forgetting. Our synthesis-driven approach addresses TTA's core limitation i.e. dependence on scarce target data, by generating high-quality, context-aware training samples that capture current environmental conditions. This enables proactive adaptation: the model updates before performance degradation occurs rather than reacting to observed failures. Among our baselines, Pseudolabel represents a standard TTA approach, while GTlabel provides an upper bound with ground-truth labels. Our results show synthesis-based adaptation substantially outperforms these approaches, particularly in data-scarce scenarios.

\mypar{Unseen species}
Our modular design naturally extends to novel or unseen species through two mechanisms. First, novel species detection can leverage out-of-distribution detection~\cite{ensemble1, mcdrop1, maxsoft1, testtime2, naff1, aff2} or open-set recognition~\cite{openset1, openset2}, together with human review~\cite{openset_human} or leveraging vision-language models (VLMs). Second, once new species data becomes available, the synthesis pipeline can integrate new animal objects, while fine-tuning can expand the classifier to new classes using class-incremental or few-shot learning. Prior work~\cite{bias,FC,BN} shows that fine-tuning limited parameters (e.g., FC, bias, or BN layers) effectively supports such updates without catastrophic forgetting.

\section{Conclusions}
\label{sec:conclusion}
This paper introduces \pjn{}, an in-situ fine-tuning system addressing domain shifts for resource-constrained IoT systems.
The effectiveness of \pjn{} in handling both spatial and temporal domain shifts opens up new possibilities for wildlife monitoring in diverse and changing environments. While we focused on wildlife classification in this work, the principles of background-aware data synthesis and drift-aware fine-tuning have broader applicability in other IoT domains where environmental conditions are dynamic and impact how we approach model deployment in resource-constrained, real-world settings.


\end{document}